\pdfoutput=1

\documentclass[11pt]{article}

\usepackage[final]{acl}

\usepackage{times}
\usepackage{latexsym}
\usepackage{amsmath}
\usepackage{hyperref}

\usepackage[T1]{fontenc}

\usepackage[utf8]{inputenc}

\usepackage{microtype}

\usepackage{inconsolata}

\usepackage{graphicx}

\title{Graph-Linguistic Fusion: Using Language Models \\ for Wikidata Vandalism Detection}

\author{Mykola Trokhymovych \\
  Pompeu Fabra University\\
  \texttt{mykola.trokhymovych@upf.edu} \\\And
  Lydia Pintscher \\
  Wikimedia Deutschland\\
  \texttt{lydia.pintscher@wikimedia.de} \\\AND
    Ricardo Baeza-Yates \\
  Pompeu Fabra University \\
  \texttt{rbaeza@acm.org} \\\And
  Diego Saez-Trumper \\
  Wikimedia Foundation  \\
  \texttt{diego@wikimedia.org} \\
}

\begin{document}
\maketitle
\begin{abstract}

We introduce a next-generation vandalism detection system for Wikidata, one of the largest open-source structured knowledge bases on the Web. Wikidata is highly complex: its items incorporate an ever-expanding universe of factual triples and multilingual texts. While edits can alter both structured and textual content, our approach converts all edits into a single space using a method we call Graph2Text. This allows for evaluating all content changes for potential vandalism using a single multilingual language model. This unified approach improves coverage and simplifies maintenance. Experiments demonstrate that our solution outperforms the current production system. Additionally, we are releasing the code under an open license along with a large dataset of various human-generated knowledge alterations, enabling further research.


\end{abstract}

\section{Introduction}
\label{sec:1_intro}
\addtocounter{footnote}{1}\footnotetext{To appear in ACL’25 (Industry Track).}
Wikidata is a large open-source, multilingual knowledge graph that plays a key role in the modern Web. It was designed as a centralized, linked repository of structured data for all Wikimedia projects, including over 300 language versions of Wikipedia~\cite{wikiedu,10.1093/llc/fqac083}. 

Beyond the Wikimedia ecosystem, Wikidata is extensively used by the most popular web services, such as search engines~\cite{kanke2021knowledge} and data for digital assistants like  \textit{Alexa} and \textit{Siri}~\cite{reagle2020wikipedia} as well as for AI models, bots, and scripts. 
Wikidata facilitates better question answering models, offers more context in search results, links to related sources efficiently, and helps reduce factual errors in large language models~\cite{wikiedu,wired_wikidata,xu-etal-2023-fine}.

\begin{figure}[b]
  \centering  \includegraphics[width=\linewidth]{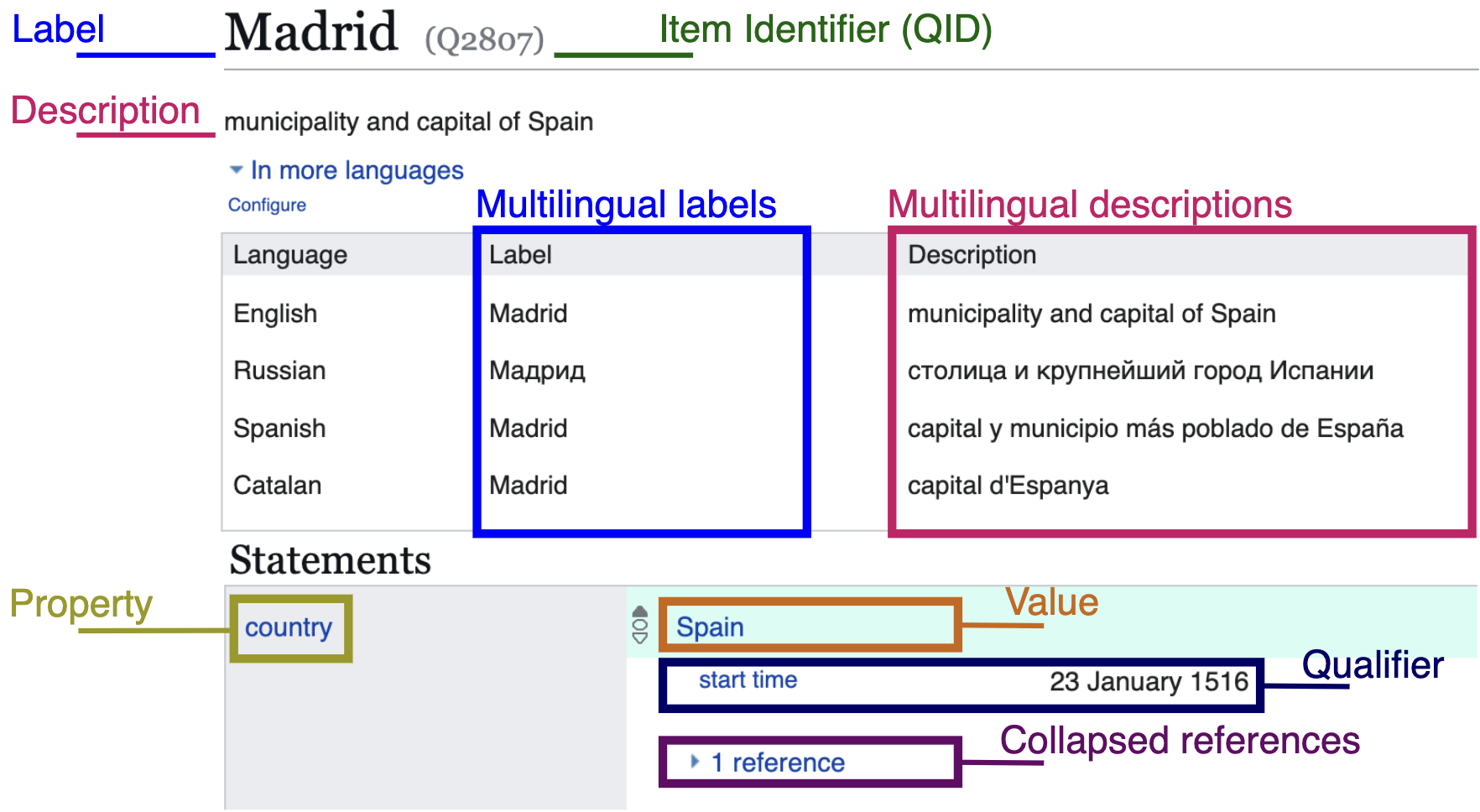}
  \caption{Diagram with the most important parts of the Wikidata record.}
  \label{fig:wikidata_record}
\end{figure}
\begin{figure}[t]
  \centering  \includegraphics[width=\linewidth]{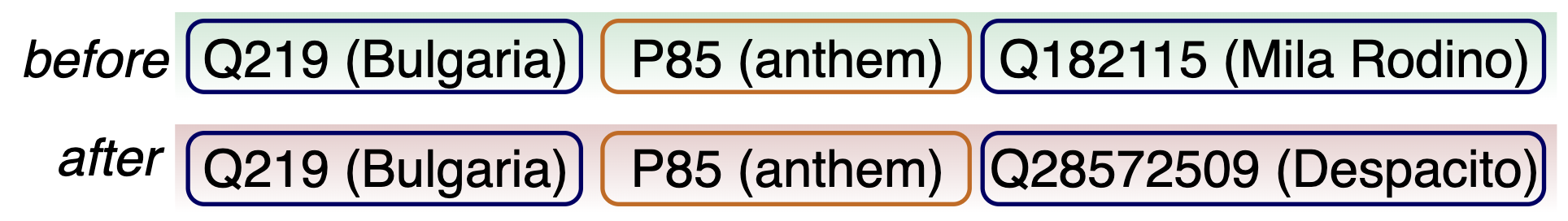}
  \caption{Example of a revision (ID: 593195479) vandalizing the Wikidata entry for Bulgaria. Original triple IDs are mapped to their corresponding English labels.}
  \label{fig:vandalism_example}
\end{figure}

Wikidata can be described as a document-oriented database focusing on items that represent any named entity~\cite{wikidata_wikipedia}. Each entity is assigned a unique identifier (ID) and can include textual information such as labels, aliases, and descriptions in multiple languages. Another essential component is the \textit{Statements}, which provides the information necessary to form semantic triples — a key component of the knowledge graph. Triples consist of tuples of \textit{\{entity,property,value\}}, where the property defines the relationship between entity and the value.
Values can be free text, numbers, dates, coordinates, or another entity. A diagram illustrating the key parts of a Wikidata record is presented in Figure~\ref{fig:wikidata_record}. 
Hence, although Wikidata provides structured relationships among entities, the building blocks of this knowledge graph include many components of unstructured data, such as multilingual descriptions or values of various types.

Given its central role in the online knowledge ecosystem, the quality of Wikidata content has relevant implications for very prominent services and products. For example, due to vandalism in Bulgaria's Wikidata Entity in 2017 (see Figure~\ref{fig:vandalism_example}), when iPhone users were asking \textit{ “What is the national
anthem of Bulgaria?,”} the answer was \textit{“Despacito”}, a popular song at that time \cite{reagle2020wikipedia}.  Vandalism has become more serious when it affects the reputation of people, institutions, or brands~\cite{saez2019online}. 
However, with Wikidata receiving around 10 edits \textit{(a.k.a revisions)} per second,\footnote{\url{https://stats.wikimedia.org/}} it becomes difficult for the human vandalism \textit{patrollers} to analyze every single edit. Therefore, several methods have been proposed to assist the community in this task by using machine learning models. In fact, in 2016, the Wikimedia Foundation developed a system named ORES that is currently supporting the vandalism detection work on Wikidata.  Unfortunately, the current ORES model is limited to certain types of edits and entities, and it cannot deal with the complexity of the different data types and topics coexisting in Wikidata. 

This paper introduces a new generation model for detecting vandalism in Wikidata that can deal with the aforementioned complexities.
A key aspect of the proposed solution is transforming all content changes, including structured data, into their textual equivalents (\textit{Graph2Text}). 
This approach allows the processing of all types of content changes by transforming them into text and using a single language model that takes advantage of the rich semantic knowledge embedded within it.


\textbf{The main contributions of this work are}: \textit{(i)}~The next-generation vandalism detection system for Wikidata, utilizing multilingual language models to improve accuracy and fairness compared to the current production model;
\textit{(ii)}~System productionalization addressing limitations imposed by resource-constrained infrastructure and product requirements;
\textit{(iii)}~The publication of a new open benchmark dataset for vandalism detection in Wikidata, containing about 5M unique samples.\footnote{\url{https://zenodo.org/records/15492678}} 


\section{Related work}
\label{sec:2_rel_work}


\subsection{Vandalism detection in Wikipedia}

Vandalism detection in Wikidata is closely related to the same problem in Wikipedia. Both services operate within the Wikimedia Foundation ecosystem, share similar editing mechanisms, and have many common users~\cite{10.1145/3041021.3053366}. Initial research on Wikipedia vandalism detection systems appeared much earlier and laid the groundwork for similar tools in Wikidata.

Early models for Wikipedia vandalism detection were binary classifiers that used generic features, such as the ratio of uppercase letters and term frequency~\cite{10.1007/978-3-540-78646-7_75}. Later studies also explored the relationship between editing behavior, editors' characteristics, link structure, and article quality on Wikipedia~\cite{article_behavior}. The most recent work proposed a vandalism detection model for Wikipedia utilizing advanced content change features based on transformer models~\cite{10.1145/3580305.3599823}.

Additionally, investigations into vandalism detection on other open-source platforms like Freebase and OpenStreetMap, which analyzed vandalism patterns and proposed various detection approaches, provide valuable insights applicable to our work due to the shared similarities among these platforms~\cite{10.1145/2556195.2556227,ijgi1030315}.

\subsection{Vandalism detection in Wikidata}

With the launch of Wikidata in 2012, it quickly became one of the most edited projects within the Wikimedia Foundation ecosystem~\cite{10.1145/2629489}. 
As with any open-knowledge project, maintaining the content reliable and verifiable has been a challenge.
The first research addressing this issue emerged, introducing WDVC-2015, a corpus designed for detecting vandalism based on the entire revision history up to that point~\cite{10.1145/2766462.2767804}. This corpus facilitated the understanding of vandalism patterns on Wikidata and provided a foundation for developing automatic vandalism detection models.

Subsequently, several approaches have been published, introducing revision classifiers to determine whether specific revisions include vandalism. These approaches employed machine learning, using features from both an edit’s content and its context~\cite{10.1145/2983323.2983740,10.1145/3041021.3053366}. One of these solutions, WDVD, proposed a model based on an extensive set of 47 content and user features, utilizing the random forest algorithm~\cite{10.1145/2983323.2983740}. Later, the Wikimedia research team introduced the ORES model, designed to function effectively in real-world applications with a much smaller feature set. This feature set was primarily established through community consultations and reflected the key concerns of Wikidata patrollers~\cite{10.1145/3041021.3053366}. 

Morover, the Wikidata Vandalism Detection Task at the WSDM Cup 2017~\cite{DBLP:journals/corr/abs-1712-05956} introduced a new dataset and received five software submissions, contributing significantly to advancements in the field~\cite{yu:2017,DBLP:journals/corr/abs-1712-06922,DBLP:journals/corr/abs-1712-06921,DBLP:journals/corr/abs-1712-06920,DBLP:journals/corr/abs-1712-06919}. 

\subsection{Bias in vandalism detection}

Even though the Wikipedia community is generally open to anyone, editors need specific skills and an understanding of community rules, which poses a challenge for newcomers. Previous research has shown that newcomer retention in Wikimedia projects is significantly affected by the reversion of their edits~\cite{article_retention,10.1145/2641580.2641614}. While newcomers and anonymous users are statistically more prone to mistakes, a biased model that unfairly cancels their edits could result in a long-term decline in the number of  editors.

One of the primary reasons for this issue is that earlier models primarily relied on user characteristics and revision metadata, using a very modest set of features to characterize actual content changes. Recent advancements in Wikipedia vandalism detection models have shown that enhancing content change processing can both improve model performance and make the system fairer for anonymous users~\cite{10.1145/3580305.3599823}.  

Similar to previous research, our focus is on processing content changes to enhance the predictive power of content features and reduce model bias. For evaluation, we employ group fairness metrics such as Disparate Impact Ratio (DIR) and the difference in AUC between privileged and unprivileged user groups~\cite{aif360}.

\section{System design}
\label{sec:3_sys_des}


\begin{figure*}[t]
  \centering
  \includegraphics[width=\linewidth]{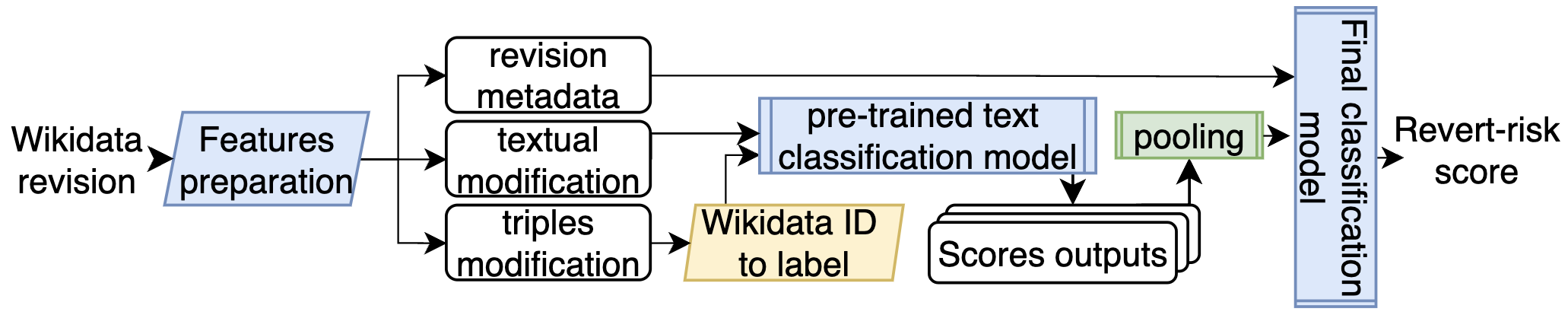}
  \caption{Wikidata vandalism detection system schema.}
  \label{fig:system_schema}
\end{figure*}


\subsection{Design requirements}
\label{sec:design}
First, our main goal is to determine if a specific Wikidata edit is vandalism. We frame this as a binary classification problem. In practice, the probability score is often more important than binary prediction, as it enables the prioritization of tasks for patrollers or the automatic reversion of changes by applying stricter thresholds.\footnote{See \url{https://www.mediawiki.org/wiki/Moderator_Tools/Automoderator} for an example from Wikipedia.}

Second, we aim to develop a single multilingual model that can process various types of content modifications (e.g., inserts, removes, changes). 
While Wikidata is largely language agnostic, it includes crucial elements like labels and descriptions that can appear in multiple languages for each record. 
Single multilingual model allows to extend the range of content edits that the model can effectively handle and reduce the infrastructure costs associated with maintaining multiple models for different content types and languages.

Third, the system requires to be efficient enough to handle a high volume of edits in a production environment. Wikidata receives about 10 edits per second, and our model should be capable of processing all of them.
We also aim to develop a system that can operate with the existing resources on the Wikimedia ML Infrastructure, called LiftWing,\footnote{LiftWing: \url{https://wikitech.wikimedia.org/wiki/Machine_Learning/LiftWing}} that
currently\footnote{As of March 1, 2025, this fact is valid.} has no GPU acceleration for inference. This high edit frequency and focus on CPU-based models rule out most LLMs.

Finally, the system must not cause undue harm to good-faith editors. Past work has shown that reverting edits by newcomers can deter new contributors~\cite{10.1145/3173574.3173929}. It is important that any deployed model does not unfairly target these newer editors.


\subsection{Architecture overview}
Our proposed system receives Wikidata revisions as input and returns a revert-risk score, indicating the probability of a given revision being reverted. The system mainly consists of three main logical steps: \textit{(i)} features preparation; \textit{(ii)} multilingual language model classifier for content processing; \textit{(iii)} final classification model to aggregate content and revision meta-features. The full system schema is presented in Figure \ref{fig:system_schema}.


\subsubsection{Feature processing} 
\label{sec:3_feat_prep}
Wikidata entity's content is represented in a complex nested structure of dictionaries and lists. Consequently, parsing content modifications can be quite challenging, as these modifications may involve structural changes (e.g., converting a single value to a list), value edits across various entities (e.g., text in different languages, numerical values, dates), and different types of content modifications (e.g., insertions, deletions, changes). Therefore, feature preparation is a critical component of the system we present. 

We distinguish three main types of features. The first type is \textit{revision metadata}, which includes features that require no additional processing and can be used directly in the final classification model (e.g., editor account creation date, time since previous revision, etc.). 

Despite Wikidata's general language-agnostic nature, its entities have textual characteristics of any language. The second feature type represents Wikidata \textit{textual modifications}, which refer to changes in elements such as entity labels, descriptions, or aliases. 
%

The third feature group is \textit{triples modification}.
Wikidata triples are composed of three parts: the entity, the property, and the value. The entity and property are represented by their corresponding Wikidata IDs. The value can also be represented by an ID, but it may also be free text, a date, a numeric value, etc. To process this content together with textual changes, we convert the triples into text by mapping the IDs to their corresponding English labels. 

\begin{figure}[t]
\centering
\includegraphics[width=\linewidth]{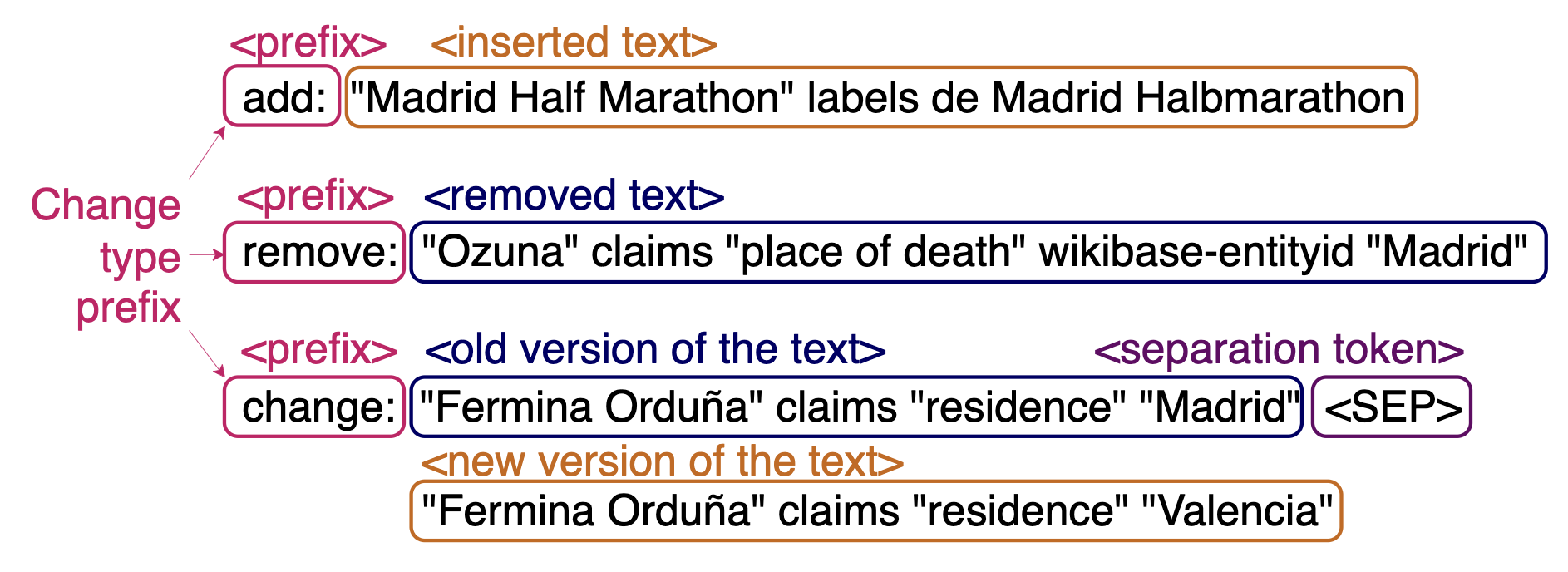}
\caption{Text processing schema.}
\label{fig:text_processing}
\end{figure}

It is important to note that both \textit{textual} and \textit{triples modification} can be of different types, such as insert, remove, and change. To process these modifications using a single language model (LM), we prepend a corresponding prefix text to the input sequence (see Figure \ref{fig:text_processing}), inspired by the "text-to-text" used in the T5 model~\cite{DBLP:journals/corr/abs-1910-10683}. This approach allows the LM to distinguish between different types of edits. 

\subsubsection{Language model classifier}
\label{sec:content_features}
To process content changes, specifically the previously discussed \textit{textual} and \textit{triples modification}, we fine-tune a single multilingual language model for binary classification tasks. 
Following the experience of a similar model for Wikipedia, we utilize the \textit{bert-base-multilingual-cased}, which was pretrained with approximately 100 languages with the largest presence on Wikipedia~\cite{10.1145/3580305.3599823,bert}.
Each revision may include multiple individual content changes of different types (e.g., a single revision might modify both a description and a factual triplet). During training, each of these changes is treated as an independent sample with the label of the revision. While inference, each of changes is independently processed by the language model classifier (LMC), with the following aggregation using mean pooling.


\subsubsection{Final classification model}
\label{sec:final_classifier}
For the final classification step, we utilize the CatBoost classifier \cite{catboost}. This model is trained using both the \textit{revision metadata} and the aggregated LMC outputs. The CatBoost classifier then generates a probability score indicating the likelihood of a revision being reverted.
Details about the hyperparameters and computational resources can be found in Appendix~\ref{sec:appendix_a}.

\subsection{Deployment details}
The complete system includes the extraction of the content using the Wikimedia API, feature engineering, and final model prediction. The inference pipeline is standardized and published under an open license in a dedicated repository of similar tools.\footnote{\url{https://gitlab.wikimedia.org/repos/research/knowledge_integrity}} 
Additional testing with editors and community discussion would still be required prior to deployment.


\section{Data preparation}
\label{sec:4_data_prep}




Initially, we collect metadata for all human-created Wikidata revisions between September 1, 2021, and September 1, 2023. 
It includes information about the Wikidata record, the user who performed the change, and specifics of the individual edit. To ensure that the revisions are human-created, we filter for revisions tagged with
\textit{Wikidata user interface}. Also, to improve data quality and reduce the noise in the revert signal, which we use as an indicator of vandalism, we additionally filter out several types of revisions (e.g., \textit{self-reverts} and revisions involved in \textit{"edit wars"}).


Wikidata entity's content is saved in the form of JSON. We extract the content for both the current and previous (parent) revisions and then compare them to identify differences.
In particular, we employ Deepdiff\footnote{\url{https://github.com/seperman/deepdiff}}
to extract the fine-grained signals from content modifications.  
We parse the content differences, getting features in the form of a list of inserts, removes, and changes. 
This includes but is not limited to alterations in descriptions, labels, and knowledge triples. Additional data processing details and explanations are included in Appendix~\ref{sec:appendix_b}.

\begin{figure}[bt]
\centering
\includegraphics[width=\linewidth]{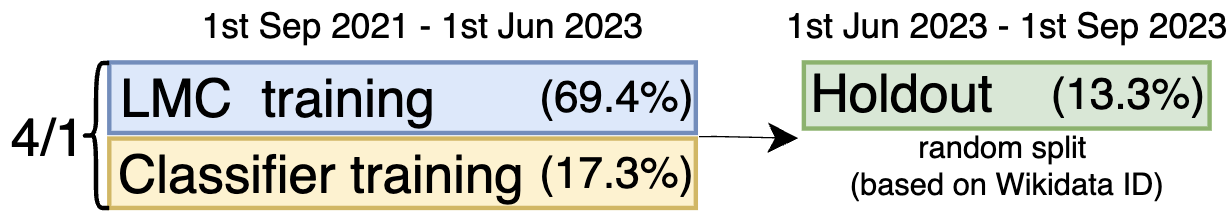}
\caption{Data splitting logic.}
\label{fig:train_test}
\end{figure}

We utilize a time-based split to allocate the last three months of collected data as the holdout testing set (see Figure~\ref{fig:train_test}).  
This portion of the dataset is reserved exclusively for the final system evaluation. 
It ensures that our evaluation strategy represents real-world usage scenarios and helps to avoid time-related anomalies.
The remaining data are used to train the components needed for the final system. 

As the proposed system consists of multiple related and independently trainable components, we divide the training part into two groups following an 80/20 split, to prevent data leakage during training.
The larger portion is used for the LMC, and the smaller for the final classifier. 
Final dataset characteristics details are presented in Appendix~\ref{sec:data_characteristics}.

\section{Evaluation}
\label{sec:5_eval}


\subsection{Baselines}

We compare our proposed model with four different baselines. As a dummy baseline, we build a \textit{Rule-based} model that considers all edits done by anonymous editors as vandalism. 
In addition, we use two strong baseline models based on subsets of the features used in the final model: the \textit{Metadata-based Classifier (MbC)} that uses only metadata features such as user group and age and the \textit{Content-based Classifier (CbC)}, ignores user characteristics and uses only content modification features. Both are classification models constructed with the same methodology described in Section~\ref{sec:final_classifier}. 

Our main reference model is ORES, the current production model for vandalism detection. This model mainly relies on metadata and includes some basic content features, such as binary indicators for changes in gender, date of birth, or English labels, to detect common vandalism patterns. We compare ORES and the previously mentioned baselines with our proposed system, \textit{Graph2Text}, which integrates advanced content modification features based on the language model, along with revision metadata features and user characteristics.

\subsection{System performance}

The primary metric we use for model comparison is the Area under the ROC curve, the AUC score. The AUC score can be interpreted as the probability that the model assigns a higher score to a random positive example than to a random negative example. Also, we compute confidence intervals (CI) for our main metric using bootstrapping (see details in Appendix~\ref{sec:appendix_c})~\cite{efron1994introduction}. 


Additionally, we employ a Filter Rate at the recall level (FR@) as suggested in previous work~\cite{10.1145/3041021.3053366}. This metric measures the proportion of edits that can be removed from Wikidata patrollers reviewing backlog, with all the remaining revisions containing a specific percentage of all vandalism. 

\begin{table}[t]
\begin{center}
\small
\caption{System performance on holdout testing set.}
{\tabcolsep=2pt
\begin{tabular}{c|c|c|c|c|c}
\hline
\label{tab:main_metrics}
\textbf{Model} & \textbf{AUC} & \textbf{CI} & \textbf{FR@99} & \textbf{FR@90} & \textbf{FR@70} \\
\hline
Rule-based & 0.760 & [0.74, 0.78] & 0.0 & 0.0 & 0.92 \\\hline
ORES & 0.859 & [0.84, 0.87] & 0.45 & 0.88 & 0.94\\\hline
MbC & 0.880 & [0.87, 0.89] & 0.55 & 0.89 & 0.94 \\\hline
CbC & 0.876 & [0.86, 0.89] & 0.60 & 0.82 & 0.93\\\hline
Graph2Text & \textbf{0.924} & [0.91, 0.93] & \textbf{0.71} & \textbf{0.91} & \textbf{0.96}
\\\hline
\end{tabular}%
}
\end{center}
\end{table}

The results of our evaluation are summarized in Table~\ref{tab:main_metrics}. Our system, \textit{Graph2Text}, significantly outperforms all other models across all metrics. 
Additionally, we observe that incorporating content features significantly improves the metrics compared to the \textit{MbC}, just as adding user features enhances the \textit{CbC}. Notably, the \textit{CbC}, which uses only advanced content features without user characteristics, performs comparably to the \textit{MbC}. This marks a significant advancement compared to previous approaches, where such performance was impossible without user characteristics. 
The performance based at FR99 indicates that with \textit{Graph2Text} (compared to ORES), patrollers will need to analyze nearly half as many revisions to detect 99\% of all vandalized samples (29\% vs. 55\%). Additional experiments, including performance evaluations for various prediction thresholds and use cases, are presented in Appendix~\ref{sec:app_performance_details}. 

\subsection{Expert evaluation}
In practice, the holdout dataset, based on community-generated data, may include revisions that have not yet been reverted or were mistakenly reverted. To enhance the validity of our evaluation, we created a subsample of 1,000 revisions for expert labeling. We divided the holdout dataset into ten bins based on scores from the ORES and \textit{Graph2Text} models separately. For each model and bin, we randomly selected fifty revisions without replacement. An experienced editor labeled these revisions as \textit{Keep}, \textit{Revert}, or \textit{Not Sure}. Revisions labeled \textit{Not Sure} were excluded from the final evaluation, resulting in 755 labeled revisions. The evaluation results, shown in Table~\ref{tab:expert_metrics}, demonstrate that consistent with the performance evaluation using community-generated labels as the ground truth, the \textit{Graph2Text} model significantly outperforms ORES on the expert-labeled data.

\begin{table}[t]
\begin{center}
\small
\caption{System performance on expert-labeled data.}
{\tabcolsep=2pt
\begin{tabular}{c|c|c|c|c|c}
\hline
\label{tab:expert_metrics}
\textbf{Model} & \textbf{AUC} & \textbf{CI} & \textbf{FR@99} & \textbf{FR90} & \textbf{FR70} \\
\hline
ORES & 0.885 & [0.879, 0.892] & 0.593 & 0.799 & 0.881 \\\hline
Graph2Text & \textbf{0.932} & [0.926, 0.937] & \textbf{0.698} & \textbf{0.846} & \textbf{0.918}
\\\hline
\end{tabular}%
}
\end{center}
\end{table}

\subsection{Fairness evaluation}

Anonymous user's edits tend to have a higher likelihood of being vandalized compared to those by registered users, primarily due to factors such as a lack of experience in editing pages or intentional identity hiding for committing vandalism. The same situation applies to newly registered users. Nevertheless, it is unacceptable for the model to discriminate based on this characteristic. On the contrary, Wikidata encourages the participation of newcomer editors.

\begin{table}[t]
\begin{center}
\small
\caption{System fairness based on holdout testing set.}
{\tabcolsep=2pt
\begin{tabular}{c|c|c|c|c}
\hline\label{tab:fair_metrics}
\textbf{Model} & \textbf{DIR\textsuperscript{anon}} & \textbf{DAUC\textsuperscript{anon}} & \textbf{DIR\textsuperscript{new}} & \textbf{DAUC\textsuperscript{new}} \\
\hline
ORES & 5.69 & 0.035  & 1.37 & -0.193\\
\hline
MbC & 4.09 & 0.097  & 1.15 & -0.155 \\
\hline
CbC & \textbf{2.07} & -0.04  & \textbf{1.08} & \textbf{-0.027}\\
\hline
Graph2Text & 4.43 & \textbf{-0.01} & 1.24 & -0.096 \\
\hline
\end{tabular}%
}
\end{center}
\end{table}


To evaluate bias against anonymous users and new editors, we use two metrics: the Disparate Impact Ratio (DIR) and the Difference in AUC score (DAUC). For more details, please refer to Appendix~\ref{sec:appendix_c}.
In particular, the closer DAUC is to 0, the better. We compare these metrics for anonymous versus registered users and newcomers versus experienced users among the registered group.
Table~\ref{tab:fair_metrics} summarizes the results of our evaluations.

Our analysis shows that our proposed model has lower DIR\textsuperscript{anon} and DIR\textsuperscript{new} values, indicating fairer treatment of anonymous and new users compared to ORES. Moreover, the difference in AUC scores between anonymous and registered users is significantly smaller, suggesting our model performs more consistently across these groups.


Although the proposed Graph2Text model demonstrates improved performance over the current ORES system, the CbC baseline, which disregards user attributes, achieves the highest fairness scores. However, our objective is to balance both predictive performance and fairness, while also maintaining applicability in scenarios where content features are not available. Consequently, we selected Graph2Text as our final model.

\section{Discussion}
\label{sec:6_conc}

To sum up, we present a study focused on developing a new generation of systems for detecting vandalism on Wikidata. 
The key innovation of the presented approach is the use of a single multilingual language model, which enables the processing of content changes in both structured and unstructured components in multiple languages. We demonstrate that the proposed system significantly outperforms the current production model in terms of both performance and fairness.

In this paper, we cover all the crucial steps needed to build a production-ready system, including the definition of design requirements, data collection and processing, feature engineering, model training, and evaluation. 

Finally, we created a new dataset capturing changes made to the Wikidata platform over a two-year period. In addition to metadata, the dataset includes detailed content edits, represented by fine-grained differences between two versions of Wikidata items. We published the dataset and the code\footnote{\url{https://github.com/trokhymovych/wikidata-vandalism-detection}} under an open license to enable further research in this area.


\subsection{Limitations} 

When interpreting the results, it's important to recognize several limitations of this study. First, the data preparation process can be improved by expanding parsing coverage, such as including changes in qualifiers or rankings. Also, using labels in non-English languages for mapping Wikidata IDs to text may enhance model performance by increasing coverage and diversifying the data.

Although the language model we fine-tuned was initially trained with about 100 languages, it still doesn't cover all of the 300+ languages represented in Wikidata. Considering these factors, we conclude that there are still issues with language diversity. Furthermore, we tested only one language model for our task. We believe that experimenting with more language models could improve the system's performance, which we leave for future research.


\subsection{Ethical considerations}

We introduce a new dataset designed to train models to predict the risk of reverts in Wikidata changes. The dataset includes metadata about revisions and editors but ensures the protection of Wikidata editors' privacy by not including any private or personally identifiable information.

We use crowd-sourced targets, which can include bias and noise, but we address this by filtering the data to minimize noise and clean the dataset. Moreover, we evaluate the system using the subsample labeled by experts. We also evaluate model fairness and ensure we reduce bias against anonymous users.  

The intended use of the model is to detect vandalism edits in Wikidata. One of the risks we care about is over-reliance on automated detection. However, the presented system includes human-in-the-loop by design, meaning human moderators retain final decision-making control while receiving enhanced assistance. 

Language models can perform differently across languages~\cite{cotterell-etal-2018-languages}. Consequently, there is a potential risk that our system may have worse performance for underrepresented languages. To address this concern, we conducted additional experiments to verify that our system significantly outperforms alternatives on both revisions with English and non-English textual content (see Appendix Section~\ref{sec:appendix_additional}).

Another potential risk of our approach is adversarial exploitation, as open access to the code and dataset could enable bad actors to design edits that bypass detection. However, we select this transparency to promote trust, accelerate further research, and enable the community to review, audit, and improve the system.

\section*{Acknowledgments}
The work of Mykola Trokhymovych is funded by MCIN/AEI /10.13039/501100011033 under the Maria de Maeztu Units of Excellence Programme (CEX2021-001195-M). This paper was partially supported by the ICT PhD program of Universitat Pompeu Fabra through a travel grant.

\bibliography{custom}

\begin{thebibliography}{32}
\providecommand{\natexlab}[1]{#1}

\bibitem[{Bellamy et~al.(2018)Bellamy, Dey, Hind, Hoffman, Houde, Kannan, Lohia, Martino, Mehta, Mojsilovic, Nagar, Ramamurthy, Richards, Saha, Sattigeri, Singh, Varshney, and Zhang}]{aif360}
Rachel K.~E. Bellamy, Kuntal Dey, Michael Hind, Samuel~C. Hoffman, Stephanie Houde, Kalapriya Kannan, Pranay Lohia, Jacquelyn Martino, Sameep Mehta, Aleksandra Mojsilovic, Seema Nagar, Karthikeyan~Natesan Ramamurthy, John Richards, Diptikalyan Saha, Prasanna Sattigeri, Moninder Singh, Kush~R. Varshney, and Yunfeng Zhang. 2018.
\newblock \href {https://arxiv.org/abs/1810.01943} {{AI Fairness} 360: An extensible toolkit for detecting, understanding, and mitigating unwanted algorithmic bias}.

\bibitem[{Cotterell et~al.(2018)Cotterell, Mielke, Eisner, and Roark}]{cotterell-etal-2018-languages}
Ryan Cotterell, Sabrina~J. Mielke, Jason Eisner, and Brian Roark. 2018.
\newblock \href {https://doi.org/10.18653/v1/N18-2085} {Are all languages equally hard to language-model?}
\newblock In \emph{Proceedings of NAACL'18: Human Language Technologies, Volume 2 (Short Papers)}, pages 536--541.

\bibitem[{Crescenzi et~al.(2017)Crescenzi, Fern{\'{a}}ndez, Calabria, Albani, Tauziet, Baravalle, and D'Ambrosio}]{DBLP:journals/corr/abs-1712-06919}
Rafael Crescenzi, Marcelo Fern{\'{a}}ndez, Federico A.~Garcia Calabria, Pablo Albani, Diego Tauziet, Adriana Baravalle, and Andr{\'{e}}s~Sebasti{\'{a}}n D'Ambrosio. 2017.
\newblock \href {https://arxiv.org/abs/1712.06919} {A production oriented approach for vandalism detection in wikidata - the buffaloberry vandalism detector at {WSDM} cup 2017}.
\newblock \emph{CoRR}, arXiv:1712.06919.

\bibitem[{Devlin et~al.(2019)Devlin, Chang, Lee, and Toutanova}]{bert}
Jacob Devlin, Ming{-}Wei Chang, Kenton Lee, and Kristina Toutanova. 2019.
\newblock \href {https://doi.org/10.18653/v1/n19-1423} {{BERT:} pre-training of deep bidirectional transformers for language understanding}.
\newblock In \emph{Proceedings of the NAACL-HLT'19}, pages 4171--4186.

\bibitem[{Dorogush et~al.(2017)Dorogush, Gulin, Gusev, Kazeev, Ostroumova, and Vorobev}]{catboost}
Anna~Veronika Dorogush, Andrey Gulin, Gleb Gusev, Nikita Kazeev, Liudmila Ostroumova, and Aleksandr Vorobev. 2017.
\newblock Fighting biases with dynamic boosting.

\bibitem[{Efron and Tibshirani(1994)}]{efron1994introduction}
Bradley Efron and R.J. Tibshirani. 1994.
\newblock \href {https://doi.org/10.1201/9780429246593} {\emph{An Introduction to the Bootstrap}}, 1st edition edition.
\newblock Chapman and Hall/CRC, New York.

\bibitem[{Grigorev(2017)}]{DBLP:journals/corr/abs-1712-06920}
Alexey Grigorev. 2017.
\newblock \href {https://arxiv.org/abs/1712.06920} {Large-scale vandalism detection with linear classifiers - the conkerberry vandalism detector at {WSDM} cup 2017}.
\newblock \emph{CoRR}, arXiv:1712.06920.

\bibitem[{Halfaker et~al.(2013)Halfaker, Geiger, Morgan, and Riedl}]{article_retention}
Aaron Halfaker, R.Stuart Geiger, Jonathan Morgan, and John Riedl. 2013.
\newblock \href {https://doi.org/10.1177/0002764212469365} {The rise and decline of an open collaboration system how wikipedia’s reaction to popularity is causing its decline}.
\newblock \emph{American Behavioral Scientist}, 57:664--688.

\bibitem[{Heindorf et~al.(2017)Heindorf, Potthast, Engels, and Stein}]{DBLP:journals/corr/abs-1712-05956}
Stefan Heindorf, Martin Potthast, Gregor Engels, and Benno Stein. 2017.
\newblock \href {https://arxiv.org/abs/1712.05956} {Overview of the wikidata vandalism detection task at {WSDM} cup 2017}.
\newblock \emph{CoRR}, arXiv:1712.05956.

\bibitem[{Heindorf et~al.(2015)Heindorf, Potthast, Stein, and Engels}]{10.1145/2766462.2767804}
Stefan Heindorf, Martin Potthast, Benno Stein, and Gregor Engels. 2015.
\newblock \href {https://doi.org/10.1145/2766462.2767804} {Towards vandalism detection in knowledge bases: Corpus construction and analysis}.
\newblock In \emph{Proceedings of SIGIR '15}, page 831–834.

\bibitem[{Heindorf et~al.(2016)Heindorf, Potthast, Stein, and Engels}]{10.1145/2983323.2983740}
Stefan Heindorf, Martin Potthast, Benno Stein, and Gregor Engels. 2016.
\newblock \href {https://doi.org/10.1145/2983323.2983740} {Vandalism detection in wikidata}.
\newblock In \emph{Proceedings of CIKM '16}, page 327–336.

\bibitem[{Kanke(2021)}]{kanke2021knowledge}
Timothy Kanke. 2021.
\newblock Knowledge curation work in wikidata wikiproject discussions.
\newblock \emph{Library hi tech}, 39(1):64--79.

\bibitem[{Kent(2019)}]{wikiedu}
Will Kent. 2019.
\newblock Why is wikidata important to you?
\newblock \url{https://wikiedu.org/blog/2019/06/03/why-is-wikidata-important-to-you/}.
\newblock Accessed on October 6, 2024.

\bibitem[{Neis et~al.(2012)Neis, Goetz, and Zipf}]{ijgi1030315}
Pascal Neis, Marcus Goetz, and Alexander Zipf. 2012.
\newblock \href {https://doi.org/10.3390/ijgi1030315} {Towards automatic vandalism detection in openstreetmap}.
\newblock \emph{ISPRS International Journal of Geo-Information}, 1(3):315--332.

\bibitem[{Potthast et~al.(2008)Potthast, Stein, and Gerling}]{10.1007/978-3-540-78646-7_75}
Martin Potthast, Benno Stein, and Robert Gerling. 2008.
\newblock Automatic vandalism detection in wikipedia.
\newblock In \emph{Advances in Information Retrieval}, pages 663--668, Berlin, Heidelberg. Springer Berlin Heidelberg.

\bibitem[{Raffel et~al.(2019)Raffel, Shazeer, Roberts, Lee, Narang, Matena, Zhou, Li, and Liu}]{DBLP:journals/corr/abs-1910-10683}
Colin Raffel, Noam Shazeer, Adam Roberts, Katherine Lee, Sharan Narang, Michael Matena, Yanqi Zhou, Wei Li, and Peter~J. Liu. 2019.
\newblock \href {https://arxiv.org/abs/1910.10683} {Exploring the limits of transfer learning with a unified text-to-text transformer}.
\newblock \emph{CoRR}, arXiv:1910.10683.

\bibitem[{Reagle and Koerner(2020)}]{reagle2020wikipedia}
Joseph Reagle and Jackie Koerner. 2020.
\newblock \href {https://doi.org/10.7551/mitpress/12366.001.0001} {\emph{Wikipedia @ 20: Stories of an Incomplete Revolution}}.
\newblock The MIT Press, Cambridge, MA.

\bibitem[{Ruprechter et~al.(2020)Ruprechter, Santos, and Helic}]{article_behavior}
Thorsten Ruprechter, Tiago Santos, and Denis Helic. 2020.
\newblock \href {https://doi.org/10.1007/s41109-020-00305-y} {Relating wikipedia article quality to edit behavior and link structure}.
\newblock \emph{Applied Network Science}, 5:61.

\bibitem[{Saez-Trumper(2019)}]{saez2019online}
Diego Saez-Trumper. 2019.
\newblock Online disinformation and the role of wikipedia.

\bibitem[{Sarabadani et~al.(2017)Sarabadani, Halfaker, and Taraborelli}]{10.1145/3041021.3053366}
Amir Sarabadani, Aaron Halfaker, and Dario Taraborelli. 2017.
\newblock \href {https://doi.org/10.1145/3041021.3053366} {Building automated vandalism detection tools for wikidata}.
\newblock In \emph{Proceedings of WWW '17 Companion}, page 1647–1654.

\bibitem[{Schneider et~al.(2014)Schneider, Gelley, and Halfaker}]{10.1145/2641580.2641614}
Jodi Schneider, Bluma~S. Gelley, and Aaron Halfaker. 2014.
\newblock \href {https://doi.org/10.1145/2641580.2641614} {Accept, decline, postpone: How newcomer productivity is reduced in english wikipedia by pre-publication review}.
\newblock In \emph{Proceedings of The International Symposium on Open Collaboration}.

\bibitem[{Simonite(2019)}]{wired_wikidata}
Tom Simonite. 2019.
\newblock Inside the alexa-friendly world of wikidata.
\newblock \url{https://www.wired.com/story/inside-the-alexa-friendly-world-of-wikidata/}.
\newblock Accessed on October 6, 2024.

\bibitem[{Tan et~al.(2014)Tan, Agichtein, Ipeirotis, and Gabrilovich}]{10.1145/2556195.2556227}
Chun~How Tan, Eugene Agichtein, Panos Ipeirotis, and Evgeniy Gabrilovich. 2014.
\newblock \href {https://doi.org/10.1145/2556195.2556227} {Trust, but verify: predicting contribution quality for knowledge base construction and curation}.
\newblock In \emph{Proceedings of WSDM '14}, page 553–562.

\bibitem[{TeBlunthuis et~al.(2018)TeBlunthuis, Shaw, and Hill}]{10.1145/3173574.3173929}
Nathan TeBlunthuis, Aaron Shaw, and Benjamin~Mako Hill. 2018.
\newblock \href {https://doi.org/10.1145/3173574.3173929} {Revisiting "the rise and decline" in a population of peer production projects}.
\newblock In \emph{Proceedings of CHI '18}, page 1–7.

\bibitem[{Trokhymovych et~al.(2023)Trokhymovych, Aslam, Chou, Baeza-Yates, and Saez-Trumper}]{10.1145/3580305.3599823}
Mykola Trokhymovych, Muniza Aslam, Ai-Jou Chou, Ricardo Baeza-Yates, and Diego Saez-Trumper. 2023.
\newblock \href {https://doi.org/10.1145/3580305.3599823} {Fair multilingual vandalism detection system for wikipedia}.
\newblock In \emph{Proceedings of the 29th ACM SIGKDD Conference on Knowledge Discovery and Data Mining}, KDD '23, page 4981–4990, New York, NY, USA. Association for Computing Machinery.

\bibitem[{Vrande\v{c}i\'{c} and Kr\"{o}tzsch(2014)}]{10.1145/2629489}
Denny Vrande\v{c}i\'{c} and Markus Kr\"{o}tzsch. 2014.
\newblock \href {https://doi.org/10.1145/2629489} {Wikidata: a free collaborative knowledgebase}.
\newblock \emph{Commun. ACM}, 57(10):78–85.

\bibitem[{Wikipedia(2024)}]{wikidata_wikipedia}
Wikipedia. 2024.
\newblock \href {https://en.wikipedia.org/wiki/Wikidata} {Wikidata}.
\newblock Accessed on October 6, 2024.

\bibitem[{Xu et~al.(2023)Xu, Liu, Culhane, Pertseva, Wu, Semnani, and Lam}]{xu-etal-2023-fine}
Silei Xu, Shicheng Liu, Theo Culhane, Elizaveta Pertseva, Meng-Hsi Wu, Sina Semnani, and Monica Lam. 2023.
\newblock \href {https://doi.org/10.18653/v1/2023.emnlp-main.353} {Fine-tuned {LLM}s know more, hallucinate less with few-shot sequence-to-sequence semantic parsing over {W}ikidata}.
\newblock In \emph{Proceedings of EMNLP'23}, pages 5778--5791.

\bibitem[{Yamazaki et~al.(2017)Yamazaki, Sasaki, Murakami, Makabe, and Iwasawa}]{DBLP:journals/corr/abs-1712-06921}
Tomoya Yamazaki, Mei Sasaki, Naoya Murakami, Takuya Makabe, and Hiroki Iwasawa. 2017.
\newblock \href {https://arxiv.org/abs/1712.06921} {Ensemble models for detecting wikidata vandalism with stacking - team honeyberry vandalism detector at {WSDM} cup 2017}.
\newblock \emph{CoRR}, arXiv:1712.06921.

\bibitem[{Yu et~al.(2017)Yu, Zhao, Wang, Xu, Shao, Wang, Ma, and Dey}]{yu:2017}
Tuo Yu, Yiran Zhao, Xiaoxiao Wang, Yiwen Xu, Huajie Shao, Yuhang Wang, Xin Ma, and Dipannita Dey. 2017.
\newblock Vandalism detection midpoint report---the riberry vandalism detector at wsdm cup 2017.
\newblock http://www.wsdm-cup-2017.org/proceedings.html.
\newblock University of Illinois at {Urbana{-}Champaign} Student Report, not published.

\bibitem[{Zhao(2022)}]{10.1093/llc/fqac083}
Fudie Zhao. 2022.
\newblock \href {https://doi.org/10.1093/llc/fqac083} {{A systematic review of Wikidata in Digital Humanities projects}}.
\newblock \emph{Digital Scholarship in the Humanities}, 38(2):852--874.

\bibitem[{Zhu et~al.(2017)Zhu, Ng, Liu, Ji, Jiang, Shen, and Gui}]{DBLP:journals/corr/abs-1712-06922}
Qi~Zhu, Hongwei Ng, Liyuan Liu, Ziwei Ji, Bingjie Jiang, Jiaming Shen, and Huan Gui. 2017.
\newblock \href {https://arxiv.org/abs/1712.06922} {Wikidata vandalism detection - the loganberry vandalism detector at {WSDM} cup 2017}.
\newblock \emph{CoRR}, arXiv:1712.06922.

\end{thebibliography}

\newpage
\appendix
\section{Modeling details}
\label{sec:appendix_a}

To process content changes we utilize the \textit{bert-base-multilingual-cased}\footnote{\url{https://huggingface.co/google-bert/bert-base-multilingual-cased}} ($\sim$178M parameters). We fine-tune the model for five epochs with an initial learning rate of $2e^{-5}$ and a weight decay of $0.01$. The batch size during training is set to 8. We reserve random 5\% of the training data as the validation set. Throughout the training process, we track the loss and select the checkpoint from the epoch where the model performs best on the validation data as the final model. Training the model requires approximately 30 GPU hours (1x AMD Radeon Pro WX 9100 16GB GPU). The choice of hyperparameter values was guided by previous approaches using similar models that have demonstrated strong performance~\cite{10.1145/3580305.3599823}.

As for the final classification model, which aggregates all the revision meta-features and outputs of LMC, we use the CatBoost classifier. We train it with 2500 iterations, a learning rate of 0.005, and a parameter selection strategy that determines the final model weights based on the iteration, achieving the best loss on the validation dataset.

\section{Data preparation}
\label{sec:appendix_b}

\subsection{Data sources}
Our dataset construction process involves extracting data from multiple sources within the Wikimedia Data Lake.\footnote{\url{https://wikitech.wikimedia.org/wiki/Analytics/Data_Lake}} 
In particular, we are utilizing the \textit{mediawiki history} table to collect metadata for all human-created Wikidata revisions and \textit{mediawiki wikitext history} table to get the Wikidata entity's content in the form of JSON. The mentioned data is available under an open license. Also, given the rarity of reverts, the initial dataset is highly imbalanced. To address this issue, we balance the dataset by retaining all reverted revisions and supplementing them with a random sample of unreverted revisions at a ratio of 1:5. The collected and processed dataset is published under an open license on the Zenodo platform to support further research.

\subsection{Data filtering}
To improve data quality and reduce the noise in the revert signal, which we use as an indicator of vandalism, we apply several filters. Specifically, we filter out \textit{self-reverts}, which are revisions reverted by the same user who created them. These reverts typically occur shortly after the revision's creation and are part of an iterative page editing process. 
Since self-reverts usually do not indicate vandalism, it is essential to filter them out to avoid falsely marking these cases as potential vandalism. 
Additionally, inspired by the process proposed in \cite{10.1145/3580305.3599823}, we filter out revisions involved in \textit{"edit wars"}. Edit wars are characterized by sequential revisions that revert one another. 
In these instances, half of the reverted revisions represent good-faith changes intended to remove vandalism. However, as it is challenging to automatically differentiate between vandalism and good-faith changes, we eliminate all such revisions to reduce noise. 
Overall, these two filters removed about 57.7\% of all revisions initially labeled as "reverted". 

\subsection{Content processing}
\begin{figure}[bt]
\centering
\includegraphics[width=\linewidth]{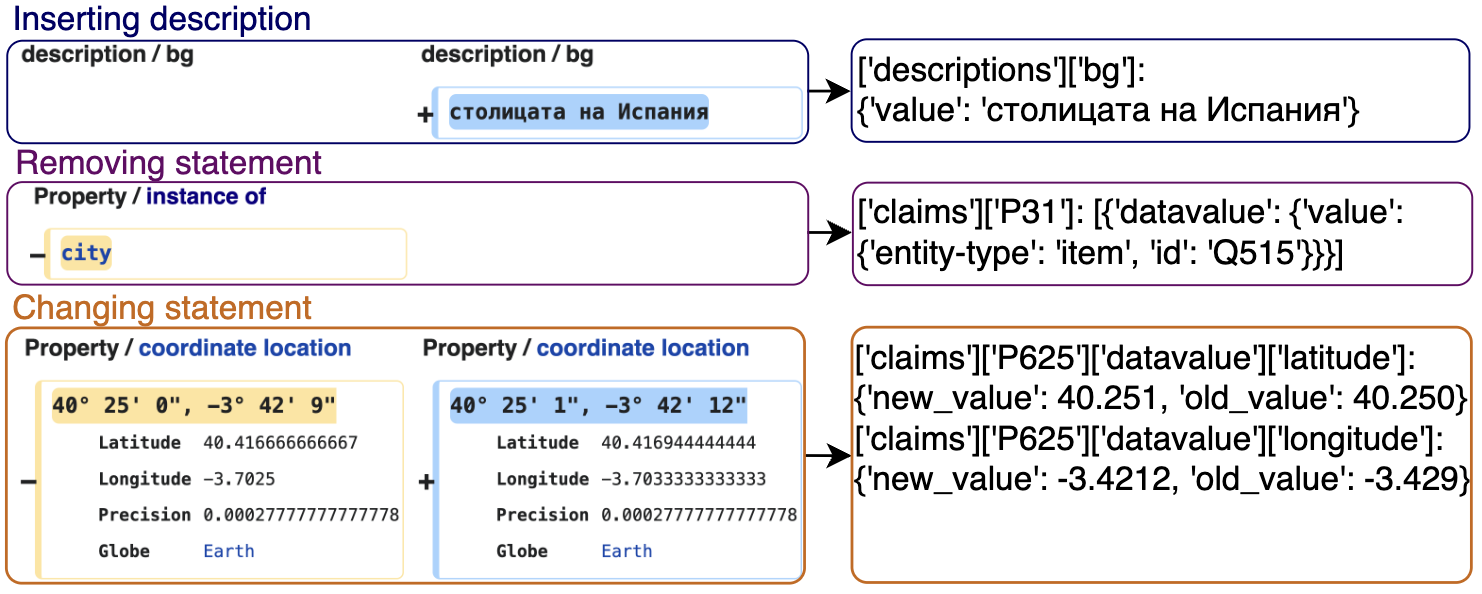}
\caption{Examples of fine-grained signals extracted from Wikidata content JSON in diverse forms and content types.}
\label{fig:example_changes}
\end{figure}

Content changes to Wikidata items include alterations in descriptions, labels, and knowledge triples (see examples in Figure~\ref{fig:example_changes}). To leverage a single language model (LM) for processing all content features, we employ specific data preparation techniques. Textual changes, such as descriptions, can be directly fed into the LM. However, graph-based features, such as knowledge triples, require additional processing. To integrate these into the LM, we convert knowledge triples into textual equivalents by mapping Wikidata IDs to their corresponding English labels. For the approximately 9\% of IDs that lack corresponding labels (\textit{i.e.} they have just an ID without a human-readable English equivalent), we map them to a default value, "unknown," which also provides a useful signal to the model. Additionally, as detailed earlier in Section~\ref{sec:3_feat_prep}, we prepend action-specific prefixes to all the input data. These prefixes supply the LM with context regarding the type of modification being processed.

\subsection{Data balancing}
We use the separate splits to train each of the components of the final system. This split is done randomly, ensuring that all revisions for a specific Wikidata entity are contained within only one of the datasets. This approach is designed to prevent contextual leakage.

Each training dataset part is further divided into separate training and validation sets. For the content model LMC, we use a random split where 5\% of the data is allocated for model validation. In contrast, for the final classifier, we employ a time-based split, mirroring the logic of the holdout set, by dedicating all revisions from the last three months for validation.

It is important to note that the obtained datasets are unbalanced. For the LMC model training, we address this imbalance by random downsampling the overrepresented class of non-reverted changes, achieving a completely balanced dataset. For the CatBoost model, we utilize the \textit{class\_weights} parameter to adjust the importance of the underrepresented class, increasing its weight according to the level of disproportion.

\section{Data characteristics}
\label{sec:data_characteristics}

The dataset is divided into two parts: a training set and a hold-out validation set, which is used for the final evaluation presented in Section~\ref{sec:5_eval}. 

The complete dataset contains 4,842,495 revisions spanning 24 months. Key data characteristics are summarized in Table~\ref{tab:data_char}. In particular, we report the rate of edits made by anonymous users and the revert rate. 

We also analyze the types of modifications made by editors (see Table~\ref{tab:data_rever}). We found that most revisions involve adding information to a Wikidata entity. This modification type also has the smallest revert rate and the lowest rate of anonymous edits. Revisions that include multiple modification types simultaneously are the most prone to containing vandalism.

It is worth noting that textual changes (modifying Wikidata entity descriptions or labels) in our dataset account for 25\% of all revisions and 16.7\% of all reverts. While English is the most popular language, it represents only 25\% of all textual changes. Other prominent languages in the top 10 include German, French, Spanish, Italian, Russian, Japanese, Swedish, Simplified Chinese, and Dutch, which, along with English, make up 62\% of the total. There are about 200 languages represented with at least 100 revisions. Revert rates vary significantly across different languages; for instance, English has a revert rate of 19\%, while Swedish has 3.7\%.

\begin{table}[tb]
\begin{center}
\small
\caption{Data characteristics.}
{\tabcolsep=3pt
\begin{tabular}{c|c|c|c|c}
\hline\label{tab:data_char}
\textbf{Dataset} & \# of samples & Period & Anon. rate & Revert rate\\
\hline
Training & 4,197,231 & 21 months &  10.7\% &  7.9\%\\
\hline
Hold-out & 645,264 & 3 months & 8.3\%   &  6.2\% \\
\hline
\end{tabular}%
}
\end{center}
\end{table}

\begin{table}[tb]
\begin{center}
\small
\caption{Revert rate by modification type.}
{\tabcolsep=3pt
\begin{tabular}{c|c|c|c}
\hline\label{tab:data_rever}
\textbf{Type} & Revert rate & \# of samples & Anon. rate\\
\hline
Insert & 11\% & 4,603,084 & 7\%  \\
\hline
Change & 29\% & 1,093,665 &  24\% \\
\hline
Remove & 35\% & 530,317 & 14\% \\
\hline
More than one type & 36\% & 183,570 & 14\% \\
\hline
\end{tabular}%
}
\end{center}
\end{table}

\section{Evaluation}
\label{sec:appendix_c}
\subsection{Confidence intervals}
To compute confidence intervals for our main metric, we employ a bootstrapping technique~\cite{efron1994introduction}. Specifically, we create 10K random samples, each of size 10K, by sampling with replacement. We then calculate the standard deviation of the AUC scores across these 10K bootstrap samples. We report the 5th and 95th percentiles for AUC as the confidence interval (CI).

\subsection{Metrics details}
For system fairness evaluation, we use the Disparate Impact Ratio (DIR). Equation~\ref{eq:DIR} presents the DIR calculation, where $Pr$ denotes the probability, $\hat{Y}$ is the predicted value, and $D$ represents a group of users. In our setup, registered users are considered the privileged group, while anonymous users and new editors are treated as the unprivileged group.

\begin{equation} \label{eq:DIR}
\frac{\operatorname{Pr}(\hat{Y}=1 \mid D=\text{unprivileged})}{\operatorname{Pr}(\hat{Y}=1 \mid D=\text{privileged})}
\end{equation}

\section{Experiments}
\label{sec:app_performance_details}
\subsection{General system performance}
\begin{figure}[bt]
\centering
\includegraphics[width=\linewidth]{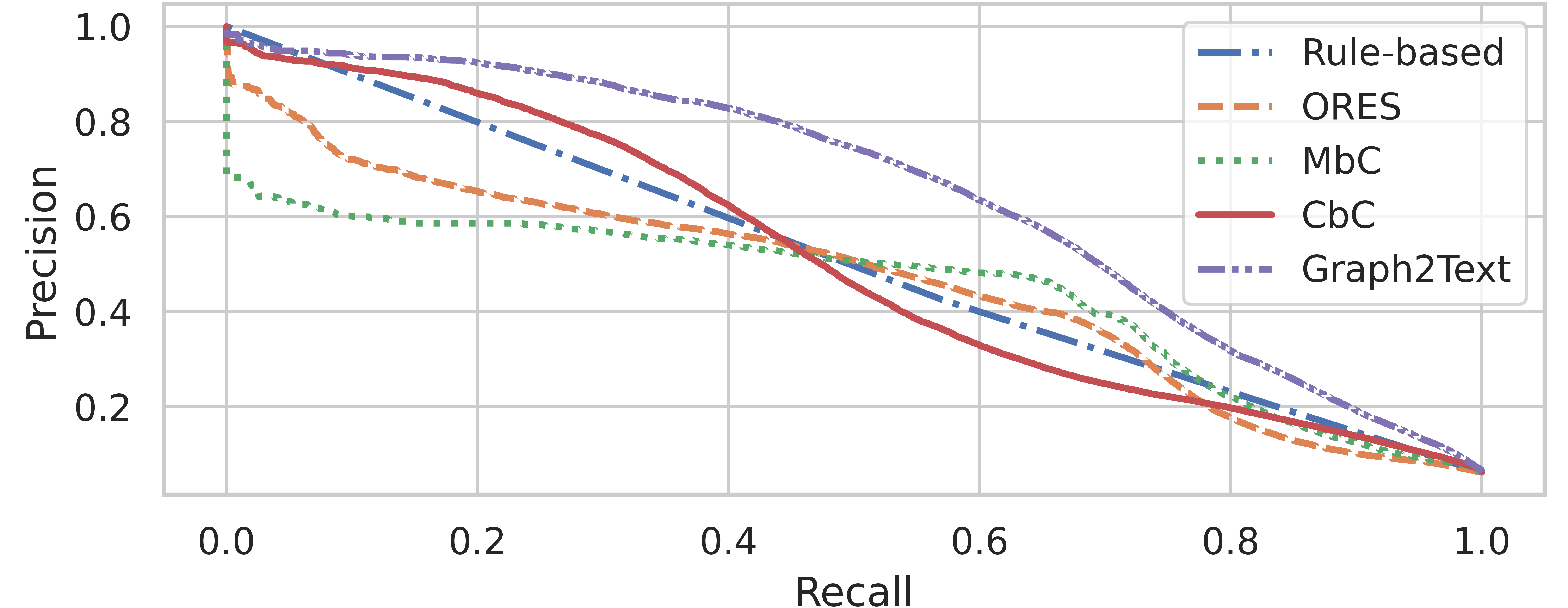}
\caption{The precision/recall curves for models.}
\label{fig:pr_plot}
\end{figure}

\begin{figure}[bt]
\centering
\includegraphics[width=\linewidth]{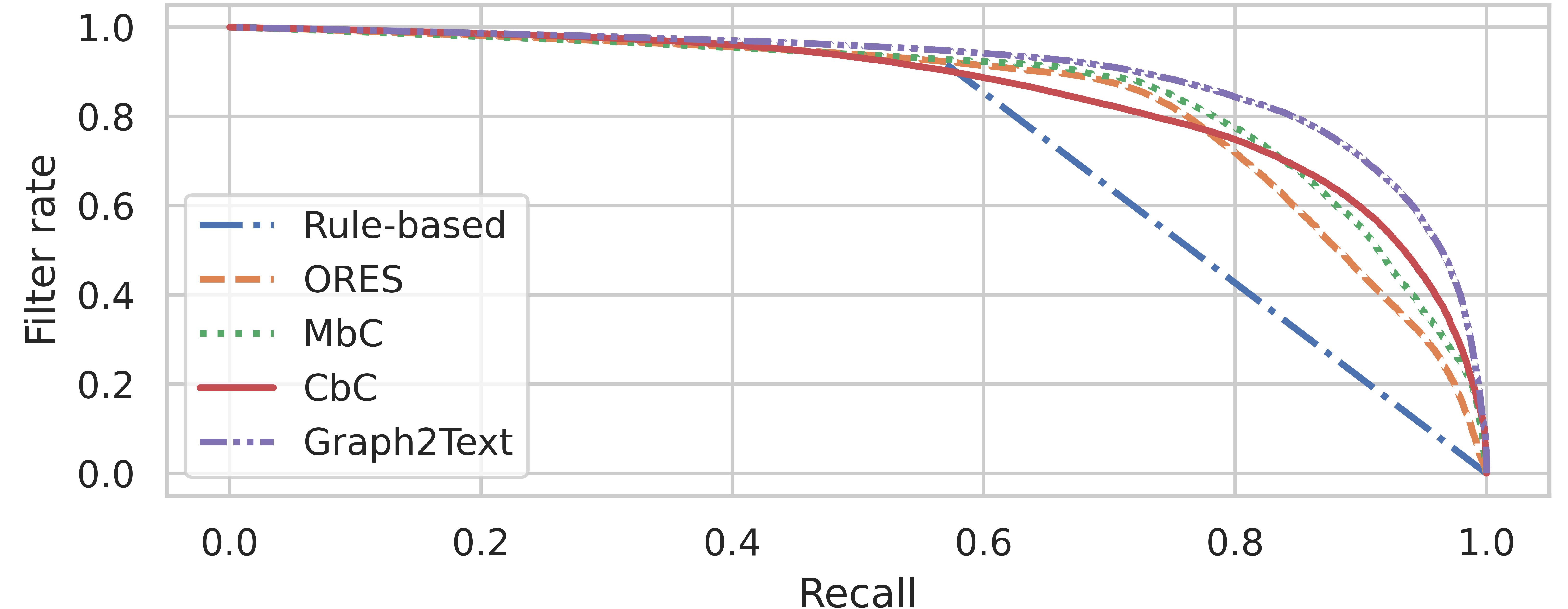}
\caption{The filter rate/recall curves for models.}
\label{fig:fr_plot}
\end{figure}

As we showed previously in Table~\ref{tab:main_metrics}, our proposed \textit{Graph2Text} system significantly outperforms all other models across all metrics. This is further confirmed by a precision/recall plot (see Figure~\ref{fig:pr_plot}), which shows that our model performs better at any threshold. We also support our analysis with a filter rate/recall plot, which highlights the dominance of the presented \textit{Graph2Text} system, especially when a high recall is needed (see Figure~\ref{fig:fr_plot}).

\subsection{Use case analysis}
\label{sec:appendix_additional}

\begin{figure}[bt]
\centering
\includegraphics[width=\linewidth]{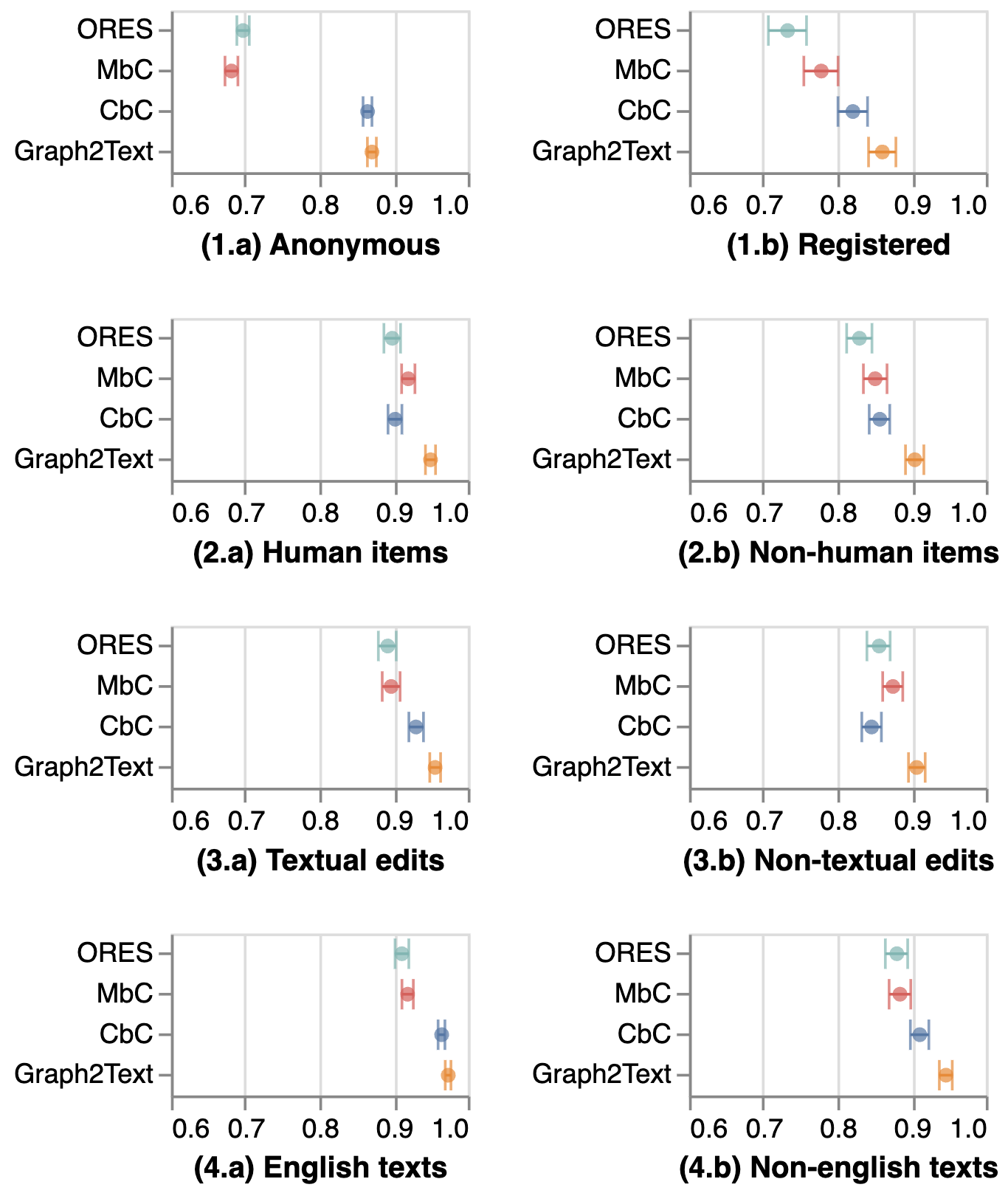}
\caption{Models performance (AUC) comparison across various Wikidata edit characteristics: (1) Edit source: (a) anonymous, (b) registered users; (2) Entity type: (a) human, (b) non-human; (3) Content type: (a) textual, (b) non-textual; (4) Textual content language: (a) English, (b) non-English.}
\label{fig:impact_plot}
\end{figure}

Additionally, we analyze how the models perform in different scenarios to understand their strengths and weaknesses and to define steps for future development and improvement.

First of all, we analyze the performance for anonymous users group. Many newcomers begin their editing as anonymous users. Retaining these new users is a priority, as they often transition into active registered editors. Failure to do so could result in a long-term decline in the number of active editors, which could significantly impact the Wikimedia environment in the future. Therefore, incorporating a bias analysis into our model evaluation is an essential step before deploying similar models in real-world contexts.

We present our findings in Figure~\ref{fig:impact_plot}. Specifically, we evaluate the model separately for anonymous and registered users. Our analysis shows that the proposed \textit{Graph2Text} system outperforms the existing ORES model for both groups. Notably, the performance difference is considerably larger for models that include content features when evaluating revisions made by anonymous users.

Wikidata contains pages about various types of entities, but pages about humans receive the most edits, accounting for about 34\% of all edits. Furthermore, modifications to human pages are more exposed to vandalism, with a 46\% higher revert rate compared to non-human pages. We compared model performance for revisions of human and non-human Wikidata entities and concluded that the proposed system outperforms the current model for both groups. Additionally, all tested systems perform better on revisions of pages about humans.

We have tested model performance on revisions with and without textual changes. As expected, even a basic content model without user features performs significantly better than the current model for handling textual edits. We also compared model performance on English and non-English textual content edits. Our findings indicate that the proposed \textit{Graph2Text} configuration is better for both groups. However, the improvement is significantly greater for English content, suggesting that the largest gains are still within English. At the same time, revisions of non-English content have over double the revert rate, and instances of vandalism persist more than twice as long for this content in Wikidata. This highlights the need to enhance vandalism detection for non-English content in the future.

\end{document}